\documentclass[runningheads]{llncs}
\usepackage[T1]{fontenc}

\usepackage{graphicx}
\usepackage{amsmath,amssymb}
\usepackage[misc]{ifsym}
\usepackage{hyperref} 
\begin{document}
\title{SegMaFormer: A Hybrid State-Space and Transformer Model for Efficient Segmentation}
%
%


%
\author{Duy D. Nguyen\inst{1, 2}, Phat T. Tran-Truong\inst{1, 2}$^{\text{(\Letter)}}$\href{https://orcid.org/0000-0003-3199-6333}{\includegraphics[scale=0.004]{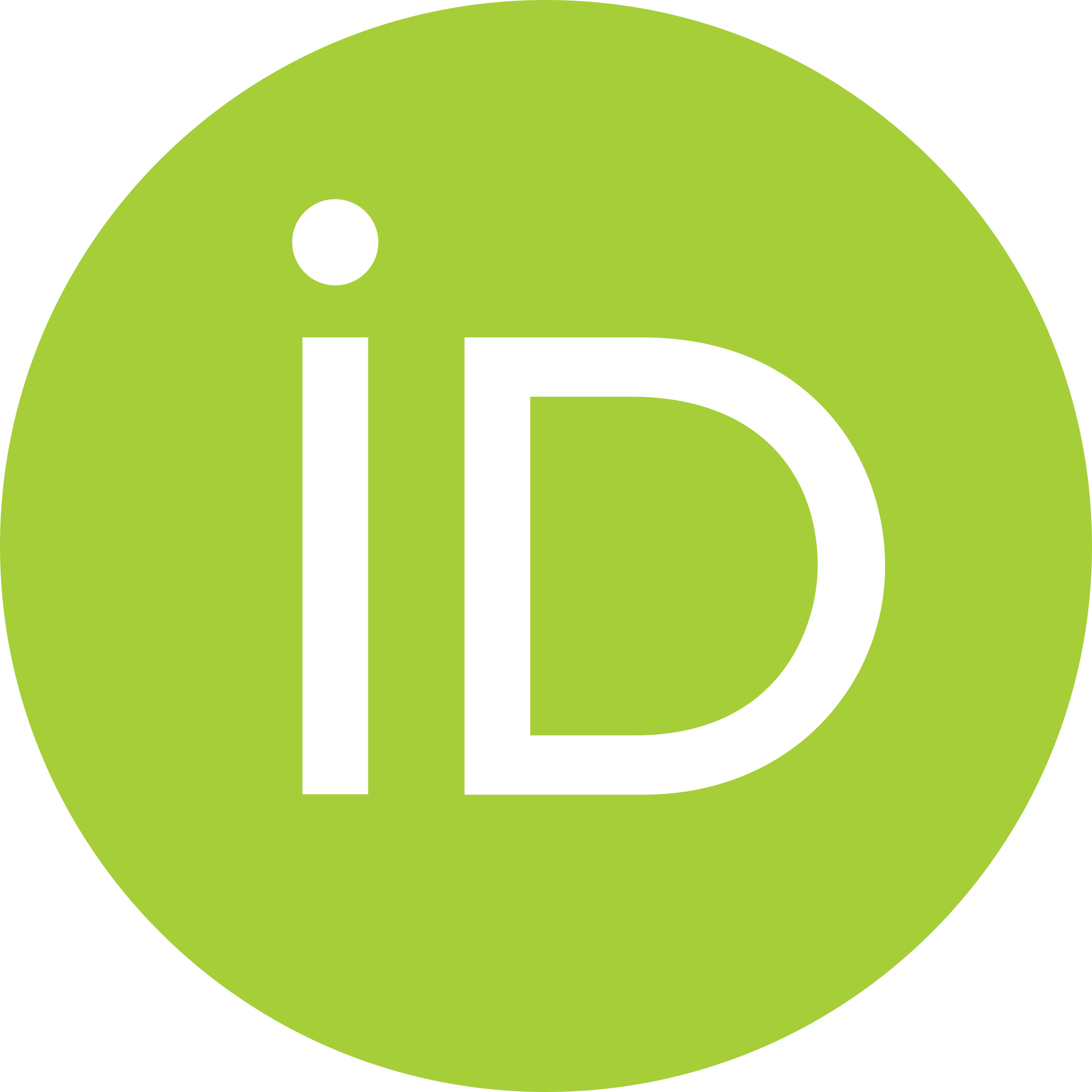}}
}

\authorrunning{Phat T. Tran-Truong}

\institute{Department of Software Engineering, Faculty of Computer Science and Engineering, Ho Chi Minh City University of Technology (HCMUT), 268 Ly Thuong Kiet Street, Dien Hong Ward, Ho Chi Minh City, Vietnam
 \and Vietnam National University Ho Chi Minh City, Linh Trung Ward, Ho Chi Minh City, Vietnam \\ ${\text{(\Letter)}}$ \email{phatttt@hcmut.edu.vn}
}

\authorrunning{Duy D. Nguyen and Phat T. Tran-Truong}
%
%
\maketitle              
\begin{abstract}
The advent of Transformer and Mamba-based architectures has significantly advanced 3D medical image segmentation by enabling global contextual modeling, a capability traditionally limited in Convolutional Neural Networks (CNNs). However, state-of-the-art Transformer models often entail substantial computational complexity and parameter counts, which is particularly prohibitive for volumetric data and further exacerbated by the limited availability of annotated medical imaging datasets. To address these limitations, this work introduces SegMaFormer, a lightweight hybrid architecture that synergizes Mamba and Transformer modules within a hierarchical volumetric encoder for efficient long-range dependency modeling. The model strategically employs Mamba-based layers in early, high-resolution stages to reduce computational overhead while capturing essential spatial context, and reserves self-attention mechanisms for later, lower-resolution stages to refine feature representation. This design is augmented with generalized rotary position embeddings to enhance spatial awareness. Despite its compact structure, SegMaFormer achieves competitive performance on three public benchmarks (Synapse, BraTS, and ACDC), matching the Dice coefficient of significantly larger models. Empirically, our approach reduces parameters by up to 75× and substantially decreases FLOPs compared to current state-of-the-art models, establishing an efficient and high-performing solution for 3D medical image segmentation.

\end{abstract}
\section{Introduction}
The proliferation of deep learning technologies is revolutionizing healthcare systems by providing robust capabilities for learning and analyzing complex patterns within medical data. A critical component of this advancement is semantic 3D volumetric segmentation, a foundational task in medical image analysis. This capability is indispensable for various clinical applications, including tumor detection and multi-organ identification, which are vital for accurate diagnosis and treatment planning.

At the dawn of the deep learning era, convolutional-neural-network-driven encoder–decoder architectures\cite{long_fully_2015,ronneberger_u-net_2015} became the dominant paradigm for medical image segmentation. Nonetheless, because their receptive fields are inherently limited due to the characteristics of CNN, these models cannot capture sufficient global context. Transformer models, utilizing the powerful attention mechanism\cite{vaswani_attention_2023}, have significantly revolutionized research in numerous fields, including computer vision and natural language processing. By adopting attention, vision models from ViT\cite{dosovitskiy_image_2021} can effectively capture long-range global dependencies through their attention layers\cite{hatamizadeh_unetr_2021,zhou_nnformer_2022}, a capability that fundamentally distinguishes them from convolutional architectures, whose operations are constrained by local inductive biases. 

To address the locality limitations of CNNs, many works integrate convolutional layers with Transformer blocks to capture global context. TransUNet \cite{chen_transunet_2021} combines a ViT encoder with a CNN decoder, while UNETR \cite{hatamizadeh_unetr_2021} applies Transformer layers directly to volumetric inputs. Subsequent models such as SwinUNETR \cite{hatamizadeh_swin_2022} and nnFormer \cite{zhou_nnformer_2022} introduce hierarchical and window-based attention for 3D data. Lighter hybrids focus attention on selected stages, as in TransBTS \cite{wang_transbts_2021} and CoTr \cite{xie_cotr_2021}, which apply attention only at coarse scales. Despite their theoretical advantages, Transformer-based models often struggle to match the generalization performance achieved by their CNN counterparts \cite{isensee_nnu-net_2024,wald_primus_2025}. Indeed, recent large-scale analyses \cite{wald_primus_2025} show that many CNN-Transformer hybrids derive most of their performance from convolutional components, and attention-heavy designs often underperform strong CNN baselines such as nnU-Net. These findings highlight architectural limitations and emphasize the need for improved positional embeddings, as demonstrated by PRIMUS \cite{wald_primus_2025}.

Although Transformers face challenges compared to CNN-based SOTA models \cite{isensee_nnu-net_2024}, their strong sequence-modeling capability makes them suited for medical image segmentation. Medical data is often multi-modal—combining scans, clinical notes, and representing images as tokens enables Transformers to integrate these heterogeneous sources within a unified model. Moreover, Transformer architectures can be computationally efficient, often requiring fewer parameters and FLOPs than CNN-based models \cite{isensee_nnu-net_2024,kuang_lw-ctrans_2025}, making them attractive for resource-constrained medical settings.

State-space models (SSMs) have emerged as efficient alternatives for long-range sequence modeling, offering \emph{linear} complexity. Structured SSMs \cite{gu_efficiently_2022} and the selective Mamba architecture \cite{gu_mamba_2024} enable hardware-efficient sequence mixing, inspiring vision variants such as VMamba \cite{liu_vmamba_2024}. In medical imaging, U-Mamba \cite{ma_u-mamba_2024} integrates Mamba blocks into a U-Net architecture to combine CNN-based local features with SSM-driven global dependencies. Nevertheless, Isensee et al.\ \cite{isensee_nnu-net_2024} report that Mamba layers alone contribute minimal gains without careful design. Swin-UMamba \cite{liu_swin-umamba_2024} addresses this by replacing Swin attention with Mamba blocks and leveraging ImageNet pretraining, achieving improvements over CNNs, ViTs, and prior Mamba variants across multiple datasets.

In many public hospitals, especially in developing regions, limited computational capacity makes large segmentation models impractical. This underscores the need for lightweight architectures that maintain high accuracy while reducing parameter counts and FLOPs, enabling reliable 3D segmentation on modest hardware. To resolve these challenges, we propose SegMaFormer, a lightweight hybrid architecture that combines the sequence-modeling efficiency of Mamba\cite{gu_mamba_2024} with the global context modeling of Transformers\cite{vaswani_attention_2023} for 3D medical image segmentation. Building upon advances in SegFormer-style tokenization \cite{xie_segformer_2021,perera_segformer3d_2024}, SegMaFormer preserves the benefits of patch-based volumetric representation while substantially reducing complexity.


In this work, we present \textbf{SegMaFormer}, a highly efficient hybrid state-space and transformer architecture for 3D medical image segmentation. Our contributions are summarized as follows:

\begin{itemize}
    \item   
    We introduce a Hybrid Transformer-Mamba encoder that combines Mamba-based state-space layers for early-stage sequence mixing with self-attention operating on compact, low-resolution tokens. This structure captures global anatomical dependencies while significantly reducing the computational burden of 3D attention, especially given that volumetric medical images expand into very long token sequences once converted into patches.

    \item 
    We enhance the baseline 3D overlapped patch embedding by integrating 3D Rotary Positional embedding (3D-RoPE), which supplies rotation-consistent positional cues while maintaining local voxel structure, boosting the performance of both Mamba and Transformer. This improves the model’s ability to capture spatial relationships in complex anatomical regions.

    \item 
    With only 2M parameters and 15 GFLOPs, SegMaFormer achieves accuracy comparable to exceeding much larger CNN and Transformer architectures, offering up to 75$\times$ fewer parameters and substantially lower computational complexity. Notably, SegMaFormer consistently surpasses the SegFormer3D\cite{perera_segformer3d_2024} baseline across all three benchmark datasets without pretraining progress, demonstrating that strong segmentation performance can be achieved without heavy model capacity.

\end{itemize}
\section{Methodology}
\begin{figure}[h!]
\includegraphics[width=\textwidth]{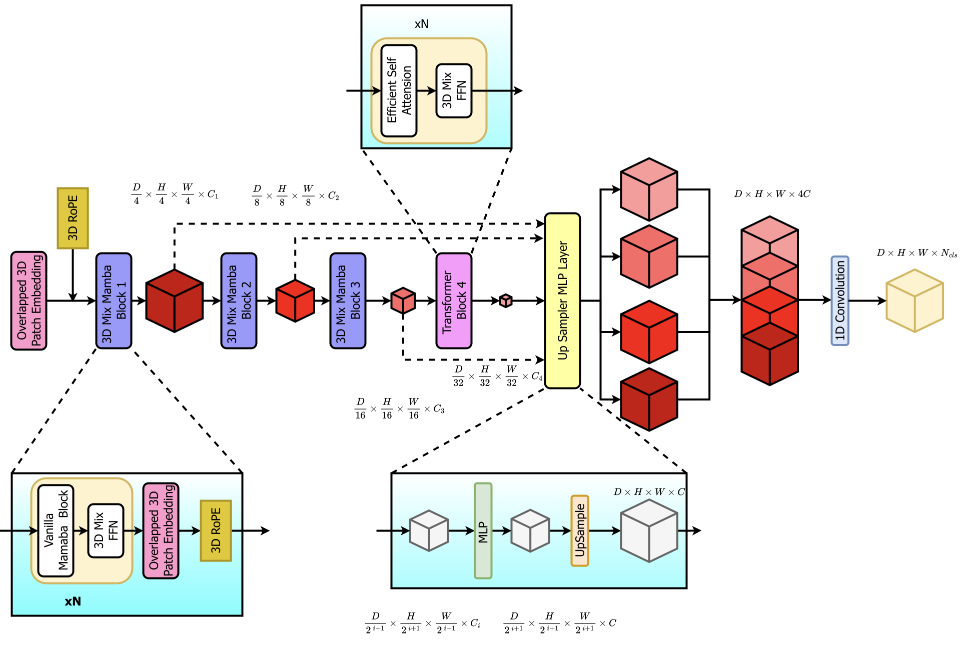}
\caption{Architecture Overview: The model input is a 3D volume 
$\mathbb{R}^{C \times D \times H \times W}$.A four-stage hierarchical Mamba-Transformer Block is adopted to derive multiscale volumetric representations. These features are then upsampled and fused by an all-MLP decoder, integrating local and global attention cues to generate the final segmentation mask.}
\label{fig1}
\end{figure}
While the attention mechanism and transformer architecture have shown a significant impact on the performance of semantic segmentation in medical images, transformer-based models have also demonstrated better computational efficiency compared to conventional CNN-based models. As well as Segformer\cite{perera_segformer3d_2024}, the base model provides an effective and lightweight framework that achieves strong segmentation performance with considerably reduced model parameters and computational complexity. However, the original Segformer baseline still suffers from quadratic complexity from the self-attention mechanism, which constrains its scalability for high-resolution or volumetric medical data and poses inference-related challenges on resource-limited or low-end computational devices.
\begin{table}
\caption{Relative comparison of model complexity in terms of parameters and GFLOPs.}
\label{tab:params_gflops}
\centering
\small
\begin{tabular}{lll}
\hline
\textbf{Architecture} & \textbf{Params(M)} & \textbf{GFLOPs} \\
\hline
nnFormer\cite{zhou_nnformer_2022} & 150.5 & 213.4 \\
TransUNet\cite{chen_transunet_2021} & 96.07 & 88.91 \\
UNETR\cite{hatamizadeh_unetr_2021} & 92.49 & 75.76 \\
SwinUNETR\cite{hatamizadeh_swin_2022} & 62.83 & 384.2 \\
Segformer3D\cite{perera_segformer3d_2024} & 4.51 & 17.5 \\
\textbf{SegMaFormer (ours)} & \textbf{2.02} & \textbf{15.2} \\
\hline
\end{tabular}

\end{table}

With the aim of alleviating this limitation, a Mamba block \cite{gu_mamba_2024} is incorporated into the original Segformer encoder to create a hybrid transformer-Mamba model, which in turn enhances computational efficiency while maintaining SOTA performance.
\paragraph{\textbf{Embedding:}}
This paper continues to use the overlapped 3D patch embedding from the baseline model. While rotary position embedding has shown clear benefits for models requiring robust sequence-dependency modeling such as Mamba and Transformers, we adopt 3D-RoPE after the embedding, following the implement idea from \cite{su_roformer_2023,gervet_act3d_2023}.
\begin{figure}[h!]
\centering
\includegraphics[width=0.4\textwidth]{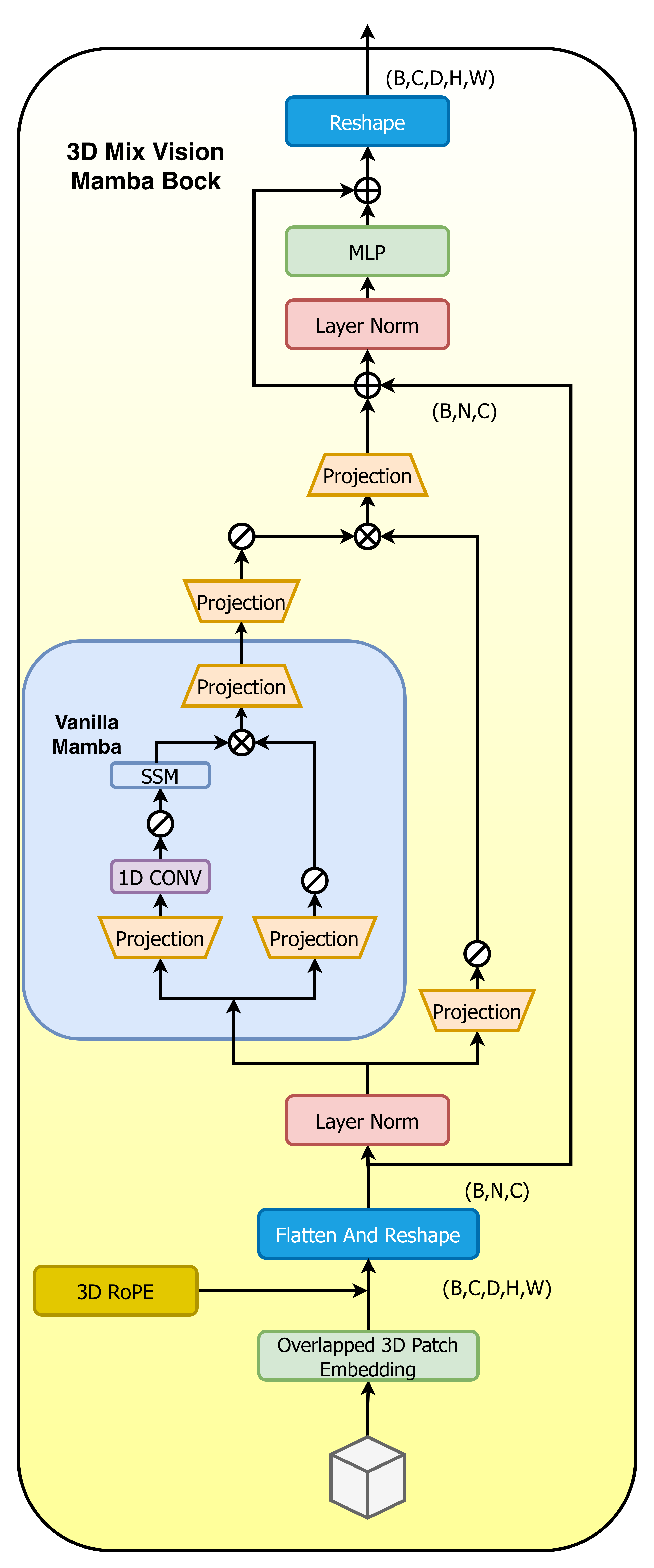}
\caption{3D Mix Vision Mamba Block with minimal reliance on convolutional layers.}
\label{fig2}
\end{figure}
\paragraph{\textbf{Encoder:}}
Processing volumetric medical data within attention-based architectures is inherently expensive because the tokenization of 3D voxel grids leads to an exponential growth in sequence length relative to spatial resolution. When a volumetric scan of resolution $D \times H \times W$ is decomposed into tokens, 
the total sequence length becomes $N = D \cdot H \cdot W$. Standard self-attention mechanisms require the computation of pairwise correlations among all tokens,
resulting in a computational complexity of 
$\mathcal{O}(N^{2} \cdot d)$,
where $d$ denotes the feature embedding dimension \cite{vaswani_attention_2023}. In 3D imaging, a moderate input resolutions (e.g., $128^{3}$ yield $N = 32{,}768$ tokens),
and thus the quadratic attention term $\mathcal{O}(N^{2} \cdot d)$ increases prohibitively large, leading to excessive memory consumption and computational latency.
To overcome the quadratic complexity of full self-attention, 
The proposed Transformer-Mamba hybrid encoder
modified the representation hierarchy by integrating 
a state-space model in the early stages 
and self-attention blocks in the last stages.

The Mamba-based state-space layers model perform sequence mixing with a computational cost of 
$\mathcal{O}\!\left(N / r \cdot d^{2}\right)$ \cite{gu_mamba_2024}, 
where $N$ is the token count, $d$ the embedding dimension, and $r$ the spatial-reduction ratio.
Otherwise, in hierarchical encoder architectures, the later stages operate on feature representations
with a smaller spatial resolution but a larger embedding dimension 
due to progressive downsampling. Under these hierarchical conditions, self-attention becomes particularly effective,
as the reduced sequence length $N$ significantly lowers its quadratic computational burden,
while the expanded embedding dimension $d$ enhances representational capacity for 
rich global context modeling.
In the case of volumetric medical images, the spatial resolution after progressive 
downsampling is often considerably smaller than the embedding dimension 
($N \ll d$). 
Consequently, applying self-attention at these deeper stages is computationally 
more efficient than in early layers and enables the model to capture 
global anatomical dependencies without incurring the prohibitive 
$\mathcal{O}(N^{2} \cdot d)$ cost associated with full-resolution attention.

This paper proposes a 3D Mamba-based block that integrates
long-range sequence modeling with efficient channel mixing.
Given a volumetric input feature map \(X\), the structure of the block is illustrated in Fig.~\ref{fig2}. Given a volumetric input feature map \(X\), the block is
formulated as
\begin{equation}
\begin{aligned}
\mathbf{X}_{\text{seq}} &= \mathrm{Flatten}(\mathbf{X}), \\[6pt]
\hat{\mathbf{X}} &= \mathrm{LayerNorm}(\mathbf{X}_{\text{seq}}), \\[6pt]
\mathbf{G} &= \mathrm{Mamba}(\hat{\mathbf{X}}), \\[6pt]
\hat{\mathbf{G}} &= \mathrm{SiLU}(\mathbf{W}_a \mathbf{G}), \\[6pt]
\mathbf{H} &= \mathrm{SiLU}(\mathbf{W}_b \hat{\mathbf{X}}), \\[6pt]
\mathbf{F} &= \mathbf{W}_p \!\left(\hat{\mathbf{G}} \odot \mathbf{H}\right), \\[6pt]
\mathbf{X}^{(1)} &= \mathbf{X}_{\text{seq}} + \mathbf{F}, \\[6pt]
\mathbf{Z} &= \mathrm{LayerNorm}(\mathbf{X}^{(1)}), \\[6pt]
\mathbf{X}^{(2)} &= \mathbf{X}^{(1)} + \mathrm{MLP}(\mathbf{Z}), \\[6pt]
\mathbf{X}_{\text{out}} &= \mathrm{Reshape}(\mathbf{X}^{(2)}).
\end{aligned}
\end{equation}
This block employs a GRU-inspired gating mechanism to dynamically regulate the information flowing through the Mamba layer, thereby preserving both data-dependent local biases and the long-range dependencies modeled by Mamba, while simultaneously accelerating convergence during training.

Traditional self-attention incurs a quadratic cost 
$\mathcal{O}(N^{2})$, which becomes prohibitive for long 3D 
sequences. Efficient attention~\cite{wang_pyramid_2021,xie_segformer_2021} reduces this cost by 
reshaping and projecting the keys,
\[
\begin{aligned}
\mathbf{\hat{K}} &= \mathrm{Reshape}\!\left(\tfrac{N}{r},\, C \cdot r\right)\mathbf{K}, \\[6pt]
\mathbf{K} &= \mathrm{Linear}\!\left(C \cdot r,\, C\right)\hat{\mathbf{K}}.
\end{aligned}
\]
lowering the complexity to $\mathcal{O}(N^{2}/r)$ in Transformer and $\mathcal{O}(N/r)$ in Mamba Block. This paper retain this formulation but apply a fixed 
reduction ratio $r=1$ across all encoder stages to evaluate the full potential of Mamba without relying on Convolution components, which can diminish the effectiveness of both Mamba and Transformer models.

\paragraph{\textbf{Decoder:}}
The decoding stage is in encoder-decoder architectures widely 
used for medical image segmentation, such as UNet 
and its variants~\cite{ronneberger_u-net_2015,isensee_extending_2022}. However, instead of the typical 
successive 3D convolutions, a lightweight decoder based on linear layers is efficient, avoids over-parameterization, 
and effectively reconstructs volumetric features.

The proposed architecture follows the baseline paradigm \cite{perera_segformer3d_2024,xie_segformer_2021}\ of fusing multi-scale 
features. As shown below, this process consists of four steps:
\begin{align}
\tilde{\mathbf{X}}_i &= \mathrm{Linear}(C_i, C)(\mathbf{X}_i), 
\quad \forall i \in \{1, \dots, 4\}, \label{eq:dec_proj} \\[4pt]
%
\mathbf{X}_i^{\text{up}} &= \mathrm{Upsample}_{s_i}(\tilde{\mathbf{X}}_i), 
\quad \forall i \in \{1, \dots, 4\}, \label{eq:dec_up} \\[4pt]
%
\mathbf{Z} &= \mathrm{Linear}(4C, C)\!\left(
\mathrm{Concat}(\mathbf{X}_1^{\text{up}}, \dots, \mathbf{X}_4^{\text{up}})
\right), \label{eq:dec_fuse} \\[4pt]
%
\mathbf{Y} &= \mathrm{Linear}(C, N_{\text{cls}})(\mathbf{Z}). \label{eq:dec_pred}
\end{align}

Features from each encoder 
stage are aggregated and projected to a unified dimensionality, similar to Unet and its variants. After 
standardization, the feature maps are upsampled, concatenated, and fused 
through a linear transformation. The resulting representation is finally 
passed through a linear prediction head (equivalent to a 3D 
$1\times1\times1$ convolution) to generate the segmentation masks. Furthermore, this work implements optional Deep Supervision (DS) auxiliary heads, similar to SOTA models. However, for tasks involving small anatomical structures, it can be observed that such auxiliary supervision does not improve performance and may even negatively impact fine-grained feature learning. Moreover, the weight of the deep supervision loss requires careful tuning, as improper balancing can lead to degraded training stability or suboptimal feature learning.

\section{Experimental Result}

Following the standard baseline model\cite{perera_segformer3d_2024}, this work adopts identical datasets and evaluation strategy to enable fair comparision between networks architectures. The proposed model is trained and assessed on three commonly used dataset benchmarks without relying on any external pretraining data. The Brain Tumor Segmentation (BraTS) \cite{antonelli_medical_2022}, Synapse Multi-Organ Segmentation (Synapse) \cite{harrigr_segmentation_2015}, and Automatic Cardiac Diagnosis (ACDC) \cite{bernard_deep_2018} datasets are evaluated in sequence.

All experiments are conducted on a dual NVIDIA RTX 4060Ti GPU using PyTorch version 2.8. We adopt the nnUnet\cite{isensee_extending_2022} framework setup for training, validation, and prediction. A weighted combination of Dice and Cross-Entropy losses is utilized to stabilize optimization and improve convergence. The learning rate is linearly warmed up from min learning rate to the initial learning rate and then decayed using cosine annealing. Training uses the AdamW optimizer\cite{loshchilov_decoupled_2019} with a base learning rate of $3\mathrm{e}{-4}$.

\subsection{Brain Tumor Segmentation (BraTs)}
This research utilizes the standard BraTS dataset from \cite{antonelli_medical_2022}, which contains 484 MRI scans of brain tumors in BraTS 2016 and 2017 Challenges from 19 hospitals, across four modals (FLAIR, T1w, T1gd, T2w). The original annotation masks contain three tumor subregions, which are edema (ED), enhancing tumor (ET), and non-enhancing tumor (NET). Following standard benchmarking practices, these labels are reorganized into whole tumor (WT), enhancing tumor (ET), and tumor core (TC) for comparison with Transformer-based methods. In this benchmark, the proposed model completely outperforms the baseline model. This demonstrates the representation-learning capability of the efficient self-attention module and the Mamba component, both of which effectively analyze the full sequence of patches.

\begin{table}
\caption{Comparison of segmentation performance across methods on the BraTS benchmark. Our model achieves competitive accuracy while using significantly fewer parameters and small FLOPS.}
\label{BCTV_brats}
\centering
\small
\resizebox{\linewidth}{!}{
\begin{tabular}{llllll}
\hline
\textbf{Methods} & \textbf{Params(M)} & \textbf{Avg(\%)} & \textbf{Whole Tumor} & \textbf{Enhancing Tumor} & \textbf{Tumor Core} \\
\hline

nnFormer~\cite{zhou_nnformer_2022} & 150.5 & 86.4 & 91.3 & 81.8 & 86.0 \\

\textbf{Ours} & \textbf{2.0} & \textbf{83.79} & \textbf{91.0} & \textbf{76.2} & \textbf{84.16} \\

SegFormer3D~\cite{perera_segformer3d_2024} & 4.5 & 82.1 & 89.9 & 74.2 & 82.2 \\

UNETR~\cite{hatamizadeh_unetr_2021} & 92.49 & 71.1 & 78.9 & 58.5 & 76.1 \\

TransBTS~\cite{wang_transbts_2021} & -- & 69.6 & 77.9 & 57.4 & 73.5 \\

CoTr~\cite{xie_cotr_2021} & 41.9 & 68.3 & 74.6 & 55.7 & 74.8 \\

CoTr w/o CNN Enc.~\cite{xie_cotr_2021} & -- & 64.4 & 71.2 & 52.3 & 69.8 \\

TransUNet~\cite{chen_transunet_2021} & 96.07 & 64.4 & 70.6 & 54.2 & 68.4 \\
\hline
\end{tabular}
}
\end{table}

\subsection{Multi-Organ CT Segmentation (Synapse)}
The Synapse dataset consists of 30 abdominal CT scans with annotations for multiple organs. Following the data split used in prior work, 18 scans are used for training and 12 for testing. Model performance is evaluated using the Dice score across eight organs, including the aorta, gallbladder, spleen, kidneys, liver, pancreas, and stomach. Table~\ref{BCTV} presents the quantitative results, demonstrating how our method compares against previous architectures in terms of accuracy, parameter efficiency. The result indices that our model achieves third-best performance, outperforming the baseline model and many larger models. High-capacity architectures such as U-Mamba \cite{ma_u-mamba_2024} and nnFormer \cite{zhou_nnformer_2022} achieve slightly higher averages, respectively, but require over 75× more parameters. Moreover, Mamba-based models, such as U-Mamba and our approach, demonstrate enhanced long-range dependency modeling, which is particularly beneficial for large-scale organs. Likewise, the Mamba-based architectures have witnessed a notable performance drop in small organs.
\begin{table}
\caption{Within the Synapse benchmark, models are ranked by their average Dice scores across all organ classes. The SegMaFormer performs exceptionally well, outperforming numerous established baselines and ranking just behind nnFormer and U-Mamba, even though these models use more than 75× the parameters and require substantially higher FLOPs.}
\label{BCTV}
\centering

\small
\resizebox{\linewidth}{!}{
\begin{tabular}{lllllllllll}
\hline
\textbf{Methods} & \textbf{Params(M)} & \textbf{Avg(\%)} & \textbf{AOR} & \textbf{LIV} & \textbf{LKID} & \textbf{RKID} & \textbf{GAL} & \textbf{PAN} & \textbf{SPL} & \textbf{STO} \\

\hline
U-Mamba~\cite{ma_u-mamba_2024} & 172.63 & 87.98 &
90.8 & 96.90 & 94.6 & 94.5 &
73.80 & 79.3 & 95.80 & 81.70 \\

nnFormer~\cite{zhou_nnformer_2022} & 150.5 & 86.57 & 92.04 & 96.84 & 86.57 & 86.25 & 70.17 & 83.35 & 90.51 & 86.83 \\

\textbf{Ours} & \textbf{2.0} & \textbf{83.33} &
\textbf{89.98} & \textbf{96.47} & \textbf{90.49} & \textbf{90.53} &
\textbf{57.29} & \textbf{70.57} & \textbf{93.03} & \textbf{78.70} \\

SegFormer3D\cite{perera_segformer3d_2024} & 4.5 & 82.15 &
90.43 & 95.68 & 86.53 & 86.13 &
55.26 & 73.06 & 89.02 & 81.12 \\

MISSFormer~\cite{huang_missformer_2021} & -- & 81.96 &
86.99 & 94.41 & 85.21 & 82.00 &
68.65 & 65.67 & 91.92 & 80.81 \\

UNETR~\cite{hatamizadeh_unetr_2021} & 92.49 & 79.56 &
89.99 & 94.46 & 85.66 & 84.80 &
60.56 & 59.25 & 87.81 & 73.99 \\

SwinUNet~\cite{cao_swin-unet_2021} & -- & 79.13 &
85.47 & 94.29 & 83.28 & 79.61 &
66.53 & 56.58 & 90.66 & 76.60 \\

LeViT-UNet-384~\cite{xu_levit-unet_2021} & 52.17 & 78.53 &
87.33 & 93.11 & 84.61 & 80.25 &
62.23 & 59.07 & 88.86 & 72.76 \\

TransClaw U-Net~\cite{chang_transclaw_2021} & -- & 78.09 &
85.87 & 94.28 & 84.83 & 79.36 &
61.38 & 57.65 & 87.74 & 73.55 \\

TransUNet~\cite{chen_transunet_2021} & 96.07 & 77.48 &
87.23 & 94.08 & 81.87 & 77.02 &
63.16 & 55.86 & 85.08 & 75.62 \\
\hline
\end{tabular}
}
\end{table}

\subsection{Automated Cardiac Diagnosis (ACDC)}
The ACDC dataset \cite{bernard_deep_2018} contains imaging data from 100 patients and is widely used to evaluate 3D segmentation methods for the left ventricle (LV), right ventricle (RV), and myocardium (Myo). Although resampling to a 1×1×1 mm isotropic resolution has been shown to be an effective preprocessing strategy validated by Wald et al. and Isensee et al. \cite{wald_primus_2025,isensee_nnu-net_2024}, we refrain from using this technique in order to ensure a fair comparison with prior benchmark studies, specially the baseline SegFormer3D \cite{perera_segformer3d_2024} model. As shown in Table~\ref{ACDC}, illustrating quantitative benchmarks of models, Mamba-based architectures typically experience a performance drop on this dataset due to their weaker inherent local inductive and spatial bias, a phenomenon also observed in recent re-evaluations of segmentation backbones \cite{isensee_nnu-net_2024}, where U-Mamba underperforms its non-Mamba counterpart. Adopting 3DRoPE into our design effectively strengthens local spatial awareness, enabling our model to recover accuracy and achieve competitive performance while keeping 2\% margin of the SOTA performance with models on average 10× higher in parameter count in computational complexity. These results further indicate that ViT-inspired tokenization strategies, which transform images into patches, when combined with appropriately designed Mamba and Transformer components, can fully replace CNNs and provide both higher efficiency and competitive accuracy.

\begin{table}
\caption{Comparison on the ACDC benchmark. Models are ranked by their average Dice scores across RV, Myo, and LV. Our method achieves competitive performance with significantly fewer parameters.}
\label{ACDC}
\centering
\small

\begin{tabular}{llllll}
\hline
\textbf{Methods} & \textbf{Params(M)} & \textbf{Avg(\%)} & \textbf{RV} & \textbf{Myo} & \textbf{LV} \\
\hline
Primus-S~\cite{wald_primus_2025} & 23.9 & 92.46 & - & - & - \\
nnFormer~\cite{zhou_nnformer_2022} & 150.5 & 92.06 & 90.94 & 89.58 & 95.65 \\

\textbf{Ours} & \textbf{2.0} & \textbf{91.11} & \textbf{90.06} & \textbf{89.1} & \textbf{94.14} \\

SegFormer3D~\cite{perera_segformer3d_2024} & 4.5 & 90.96 & 88.5 & 88.86 & 95.53 \\

LeViT-UNet-384~\cite{xu_levit-unet_2021} & 52.17 & 90.32 & 89.55 & 87.64 & 93.76 \\

SwinUNet~\cite{cao_swin-unet_2021} & -- & 90.00 & 88.55 & 85.62 & 95.83 \\

TransUNet~\cite{chen_transunet_2021} & 96.07 & 89.71 & 88.86 & 84.54 & 95.73 \\

UNETR~\cite{hatamizadeh_unetr_2021}& 92.49 & 88.61 & 85.29 & 86.52 & 94.02 \\

R50-VIT-CUP~\cite{chen_transunet_2021} & 86.00 & 87.57 & 86.07 & 81.88 & 94.75 \\

VIT-CUP~\cite{chen_transunet_2021} & 86.00 & 81.45 & 81.46 & 70.71 & 92.18 \\
\hline
\end{tabular}

\end{table}

\section{Limitation and Conclusion}
In this work, an efficient lightweight model was proposed to address the growing computational and generalization challenges of 3D medical image segmentation. By integrating a 3D-RoPE in patch merging embedding, Mamba-based state-space layers in the early high-resolution stages, and applying self-attention only at deeper, low-resolution scales, the proposed framework achieves an effective balance between accuracy and complexity. This design enables substantial reductions in parameters and FLOPs while maintaining competitive performance across multiple benchmarks, demonstrating that compact architectures can remain highly effective even in data-limited medical imaging scenarios. Furthermore, these findings suggest that the parameter number requires careful consideration, as over-parameterization does not lead to significant performance gains. Despite its promising results, the model still presents opportunities for improvement. The performance on small organs and sharp anatomical boundaries indicates room for further refinement, particularly in capturing fine-grained spatial details under limited training data. The proposed approach leverages lightweight hybrid architectures and provides a practical foundation for scalable, resource-efficient, and clinically deployable 3D medical image segmentation.

\section*{Acknowledgement}
We acknowledge the support of time and facilities from Ho Chi Minh City University of Technology (HCMUT), VNU-HCM for this study. 
%
%
%
%
\bibliographystyle{splncs04}
\bibliography{references}

\end{document}